%% file: main.tex
\documentclass{article}

\usepackage[final]{corl_2026} % camera-ready, non-anonymous

\input{math_commands.tex}

\usepackage{graphicx}
\usepackage{booktabs}
\usepackage{multirow}
\usepackage{array}
\usepackage{xcolor}
\usepackage{amsmath}
\usepackage{amssymb}

\newcommand{\affilmark}[1]{\textsuperscript{#1}}

\newcommand{\advisemark}{\textsuperscript{\ensuremath{\dagger}}}


\title{When Video Misreads:\\
Closed-Loop Distillation of Reading Heuristics for\\
Exploratory Manipulation Trace QA}

\author{
  Haizhou Ge\affilmark{1} \quad Yufei Jia\affilmark{1} \quad Yue Li\affilmark{2} \quad Zhixing Chen\affilmark{1} \quad Lu Shi\affilmark{1}\\
  \bf Lei Han\affilmark{2} \quad Guyue Zhou\affilmark{1} \quad Ruqi Huang\affilmark{1}\advisemark\\[0.5em]
  \mdseries \affilmark{1}Tsinghua University \qquad \affilmark{2}DISCOVER Robotics\\[0.3em]
  \advisemark Advising.
}

\begin{document}
\maketitle

\begin{abstract}
    \input{sections/0_abstract.tex}
\end{abstract}


\input{figures/ai_generated/latex_include.tex}

\section{Introduction}
\label{sec:intro}
\input{sections/1_introduction.tex}

\section{Related Work}
\label{sec:related}
\input{sections/2_related_work.tex}

\section{Closed-Loop Trace Distillation}
\label{sec:method}
\input{sections/3_method.tex}

\section{Main Results}
\label{sec:main_results}
\input{sections/4_main_results.tex}

\section{Modality Ablation}
\label{sec:ablations}
\input{sections/5_ablations.tex}

% CoRL requires \section{Limitations}; counts toward the 8-page budget.
\section{Limitations}
\label{sec:limitations}
\input{sections/6_limitations.tex}

\section{Conclusion}
\label{sec:conclusion}
\input{sections/7_conclusion.tex}

\clearpage
\acknowledgments{If accepted, the camera-ready version will include acknowledgments to reviewers, collaborators, and funding agencies.}

\bibliography{references}

\appendix
\input{sections/A_appendix.tex}

\end{document}

%% file: math_commands.tex
\providecommand{\eg}{\emph{e.g.,}}

%% file: sections/0_abstract.tex
% Abstract: make the task, artifact, and inference boundary explicit.
Exploratory manipulation often turns an apparent failed attempt into the key evidence for what to do next. For example, a robot pulls a locked cabinet drawer, fails, and only succeeds after opening the lock. The failed pull reveals a latent precondition (the drawer is locked) that determines the minimal-success action chain (the fewest actions that complete the task), here [lock-open, drawer-pull]. Correctly reading this trace is therefore the prerequisite for recovering that chain. We formalize this setting as \emph{Exploratory Manipulation Trace QA} (EMT-QA): given synchronized video and proprioception from an exploratory trace, predict the minimal-success action chain under the latent precondition revealed by the probe. However, even state-of-the-art VLMs and embodied multimodal LLMs misread this evidence: they do not reliably recover the chain from raw video, raw proprioception, or their combination.
We introduce \emph{Closed-Loop Trace Distillation}, a pipeline that uses a per-task coding agent to inspect labeled training traces and distill a one-line natural-language prompt over the trace, which we call the \emph{Distilled Reading Heuristic} (DRH). At inference, no agent is invoked and no model weights are updated; a frozen VLM receives the raw trace plus the DRH as a prompt entry. Across three simulator and two real-robot tasks, the DRH improves chain accuracy by $+0.38$ to $+0.47$ over the best raw-modality baseline. The same DRH also serves as the sole specification for one-shot programmatic classifiers that match the prompted VLM.

%% file: figures/ai_generated/latex_include.tex
\begin{figure*}[t]
    \centering
    \includegraphics[width=0.99\textwidth]{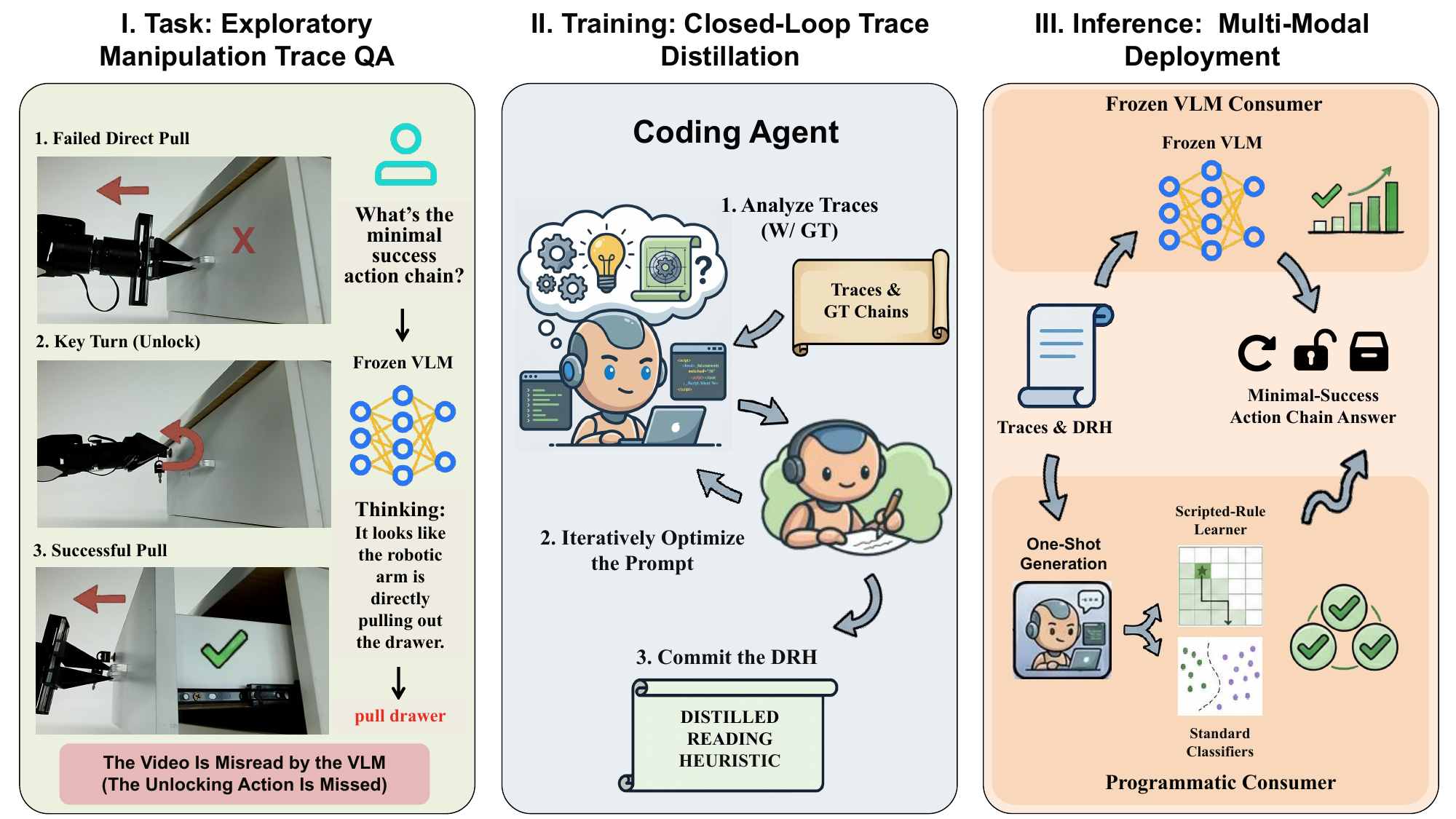}
    \caption{\textbf{Closed-Loop Trace Distillation.} (1) EMT-QA: given an exploratory manipulation trace pairing video with a proprioceptive trajectory, predict the minimal-success action chain (e.g., [CCW turn-knob, pull-door]); current VLMs misread the trace and fail to recover the chain from raw modalities. (2) A closed-loop coding agent iterates on training traces, discovers how to interpret the trace, and distills the finding into a one-line Distilled Reading Heuristic (DRH). (3) The DRH serves two classes of consumer: a frozen base VLM that reads it as a prompt entry ($+0.38$ to $+0.47$ chain-accuracy lift), and programmatic classifiers generated one-shot from the same heuristic.}
    \label{fig:teaser}
\end{figure*}

%% file: sections/1_introduction.tex
% Introduction (clarified task-method arc)

When a person opens an unfamiliar drawer or door, they almost always try the simplest action first, feel resistance, then commit to the correct one (Figure~\ref{fig:teaser}). A drawer pull that fails until the cabinet is unlocked, or a clockwise knob turn that stalls before a counter-clockwise success, reveals a latent precondition that determines which subsequent action chain can succeed~\citep{wang2025adamanip,xiang2020sapien,mo2021where2act,geng2023gapartnet}. We study this setting as \emph{Exploratory Manipulation Trace QA} (EMT-QA): given synchronized video and proprioception from the full exploratory trace, a model must choose the minimal-success action chain, the fewest actions that complete the task (\eg{} ``counter-clockwise turn-knob, pull-door''), implied by the probe.

Existing work on robot manipulation traces falls into several lines, each with a different emphasis: failure-detection framings such as AHA~\citep{duan2024visionlanguagemodel} and Guardian~\citep{pacaud2025scaling} tag the probe as a defect; visible-object trajectory taggers~\citep{galoaa2026motiono,schroeder2025rover} read object motion or recursively decompose video without exposing the hidden proprioceptive discriminator; trained VLA/VLM stacks~\citep{wang2026think,zitkovich2023rt2,kim2024openvla,driess2023palme} merge proprioception tokens into policies without exposing how they read the exploratory trace; and robotics-VQA datasets~\citep{chen2025robovlm} benchmark generic manipulation QA. None of these directly targets the EMT-QA chain-prediction structure. General-purpose VLMs and embodied multimodal LLMs are the natural fallback, since they can in principle read both the video and the proprioceptive trajectory. In practice they recover the cue from neither: a human observer can pick up the discriminative signal (a small change in the target end-effector orientation, a gripper open/close delta, or a phase-split in the trajectory) from the raw video, while current models cannot, and they also fail to parse it directly from the proprioceptive trajectory. In our experiments, even a recent embodied multimodal LLM with native video + proprioception collapses to $1/21$ on a safe-opening task. These failures motivate an explicit artifact that instructs a downstream model how to read the exploratory trace.

We propose \emph{Closed-Loop Trace Distillation}, a pipeline that converts labeled EMT-QA traces into a one-line natural-language prompt, which we call the \emph{Distilled Reading Heuristic} (DRH). During training, a per-task coding agent inspects the trace input, proposes candidate heuristics over the trace, and commits a candidate only after it passes held-out gates on \emph{chain accuracy} (the fraction of episodes whose predicted chain matches ground truth). During inference, the agent disappears: a frozen general-purpose VLM receives the raw trace, the chain prompt, and the DRH as a prompt entry, then emits the chain answer with no tool calls and no weight updates. We call this consumer the \emph{Distilled-Prompt VLM}; the same frozen model without the DRH is the \emph{Naked-Modality VLM}. Across three simulator and two real-robot tasks, the DRH lifts chain accuracy by $+0.38$ to $+0.47$ over the best raw-modality baseline. The same DRH also serves as the sole specification from which a coding agent one-shot generates programmatic baselines that match the prompted VLM on held-out splits.

Overall, our contributions are:
\begin{enumerate}
    \item We formalize \emph{Exploratory Manipulation Trace QA} (EMT-QA), a multimodal QA task in which the probe segment of an exploratory trace reveals the latent precondition that determines the minimal-success action chain (\S\ref{sec:method}).
    \item We propose \emph{Closed-Loop Trace Distillation}, a training pipeline that uses a per-task coding agent to distill labeled traces into a \emph{Distilled Reading Heuristic} (DRH), with no model-weight updates and no agent calls at inference; across five tasks the DRH lifts a frozen base VLM's chain accuracy by $+0.38$ to $+0.47$ over the best raw-modality baseline (\S\ref{sec:method},~\S\ref{sec:main_results}).
    \item We further show that the same DRH transfers as the sole specification for a coding agent to one-shot generate programmatic baselines that match the prompted VLM (\S\ref{sec:supervised_baselines}).
    \item A cross-model modality ablation confirms that neither a frozen general-purpose VLM nor a recent embodied multimodal LLM recovers the lift from raw video, raw proprioception, or their combination, locating the contribution in the DRH rather than the underlying modality stack (\S\ref{sec:ablations}).
\end{enumerate}

%% file: sections/2_related_work.tex
% Related Work (compressed)

\paragraph{VLM/VLA Models for Manipulation.}
Four lines of work read or train on robot manipulation traces but emit different predictions: Robo2VLM~\citep{chen2025robovlm} (benchmark VQA from large-scale robot data), Motion-o~\citep{galoaa2026motiono} (visible object-motion tags), ROVER~\citep{schroeder2025rover} (recursive video decomposition for task-progress), and Think Proprioceptively~\citep{wang2026think} (VLA with proprioception tokens for next-action policies). In contrast, we frame chain prediction over an exploratory trace with the base VLM frozen and the answer readable from a one-line natural-language heuristic over the trace.

\paragraph{Failure Reasoning Systems.}
AHA~\citep{duan2024visionlanguagemodel} and Guardian~\citep{pacaud2025scaling} are representative failure-reasoning approaches on manipulation traces; they treat a failed first attempt as a defect to detect or discard. Our framing is opposite: a failed first attempt signals that a latent precondition is unsatisfied, and the corrective sub-trace inserted before the successful retry is positive evidence rather than a defect.

\paragraph{Multimodal-Planning and Robotics-VQA Benchmarks.}
RoboVQA~\citep{sermanet2023robovqa} (long-horizon robotics VQA), ManipBench~\citep{zhao2025manipbench} (multiple-choice low-level manipulation), and EgoPlan-Bench2~\citep{qiu2024egoplanbench} (egocentric MLLM planning) target related domains. EMT-QA adopts their multimodal multiple-choice protocol but differs on two axes. (i) The QA task itself: we ask for a minimal-success chain over an exploratory trace whose probe segment reveals a latent precondition, rather than long-horizon planning, low-level manipulation reasoning, or egocentric plan steps. (ii) The evaluation granularity: we report per-truth-chain accuracy in addition to aggregate, exposing ties where models default to guessing from the candidate options that aggregate-only scoring hides.

\paragraph{Coding-Agent Orchestration and Agent Loops.}
CodeAct~\citep{wang2024executable} and ROSBag MCP Server~\citep{fu2025rosbag} motivate the script-writing reasoner and log-inspection interface; the Claude Code design space~\citep{liu2026dive} supplies the iteration / context / permissions vocabulary that our closed-loop training loop inherits. Our iteration agent's reason-act-revise template follows the ReAct~\citep{yao2023react}, Reflexion~\citep{shinn2023reflexion}, and Self-Refine~\citep{madaan2023selfrefine} pattern, applied to a one-line natural-language prompt rather than to free-form LLM output. The downstream-VLM consumption of the DRH at inference is an in-context-learning conditioning step in the sense of \citet{brown2020gpt3} with the DRH acting as a chain-of-thought-style trigger~\citep{wei2022cot}.

\paragraph{Code-as-Policy and Generalist Embodied Baselines.}
Code-as-policy systems (Code as Policies~\citep{liang2023codeaspolicies}, ProgPrompt~\citep{singh2023progprompt}, VoxPoser~\citep{huang2023voxposer}, and RoboScript~\citep{chen2024roboscript}) invoke an LLM at inference to synthesize control programs. Generalist embodied baselines (Gemini Robotics~\citep{team2025gemini} and the embodied multimodal LLM in \S\ref{sec:ablations}) are strong monolithic models we do not retrain. In contrast, our closed-loop step trains no weights; it edits a one-line reading heuristic the base VLM cannot infer on its own, the same artifact doubles as a specification for the programmatic baselines in \S\ref{sec:supervised_baselines}, and the base VLM emits only a discrete chain label without invoking a coding agent at inference.

%% file: sections/3_method.tex
% Method (aggressively compressed; details deferred to appendix)

In this section, we describe \emph{Closed-Loop Trace Distillation}, a pipeline that trains, per task, a \emph{Distilled Reading Heuristic} (DRH) through a closed-loop coding agent against ground-truth chain labels. The DRH is the only trained artifact, consumed at inference as a prompt entry by a downstream model (a frozen base VLM in \S\ref{sec:main_results}, a coding-agent-generated programmatic baseline in \S\ref{sec:supervised_baselines}); no coding agent is invoked at inference and no model weights are updated.

\subsection{Problem setup}
\label{sec:setup}

We introduce \emph{Exploratory Manipulation Trace QA} (EMT-QA), a chain-prediction task over an \emph{exploratory manipulation trace}. The input is a synchronized stream of (i) a manipulation video; (ii) a proprioceptive trajectory of the target end-effector pose (3-D rotation + 3-D position) and a 1-D gripper open/close state; and (iii) a fixed chain prompt (a task description plus a multiple-choice answer list of candidate chains; Appendix~\ref{app:prompts}). The output is the \emph{minimal-success action chain}, the fewest actions that complete the task, implied by the trace (\eg{} [CCW turn-knob, pull-door] for safe, [lock-open, drawer-pull] for cabinet). For brevity, we use \emph{trace} for the synchronized (i)+(ii) stream and \emph{chain answer} for the minimal-success action chain; the metric is \emph{chain accuracy}, the fraction of episodes whose predicted chain answer matches ground truth. Representative video frames for the five paper-scope tasks are in Figure~\ref{fig:task_gallery}.

\subsection{Closed-loop distillation pipeline}
\label{sec:pipeline}

\begin{figure}[t]
    \centering
    \includegraphics[width=0.99\linewidth]{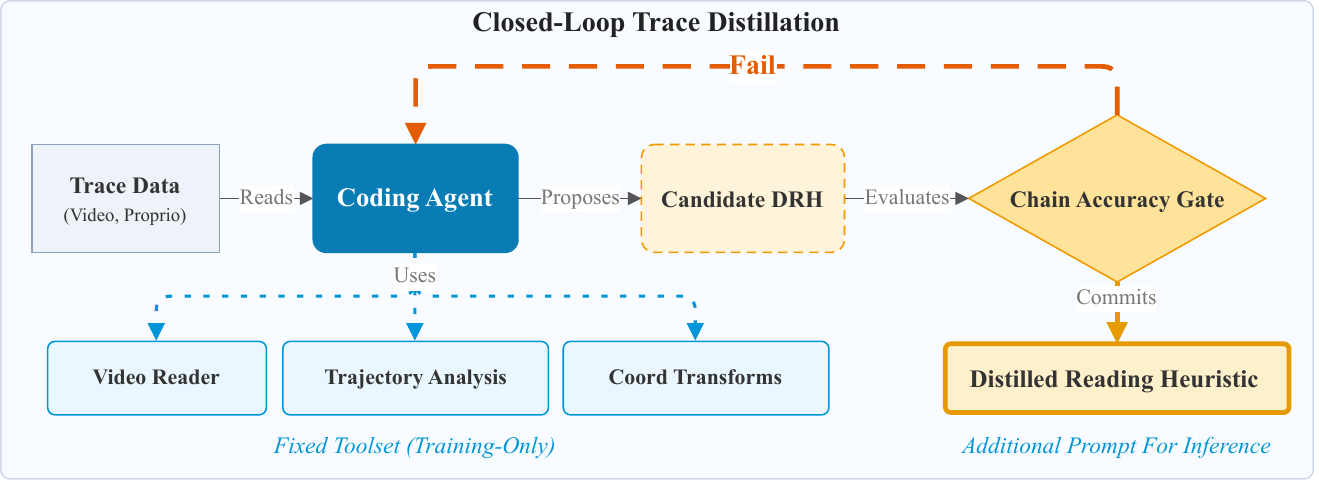}
    \caption{Closed-Loop Trace Distillation training loop. A coding agent iterates on the task's traces, probes the trace input (video + proprioceptive trajectory) with a fixed toolset, proposes a candidate Distilled Reading Heuristic (DRH), and evaluates it on held-out groups; the candidate is committed only when consecutive training groups clear a chain-accuracy gate, otherwise the agent re-iterates (fail loopback). The committed DRH is the only trained artifact (no model weights are updated).}
    \label{fig:pipeline}
\end{figure}

Figure~\ref{fig:pipeline} shows the training loop.

\textbf{Iteration agent.} Per task, an iteration agent (Claude Code~\citep{liu2026dive}) probes the trace input with a fixed toolset (video reader, trajectory analysis library, and coordinate transforms), seeing only training traces and their ground-truth chain labels; the held-out fraction is reserved for evaluation. The agent proposes a candidate DRH and commits it only when it clears a held-out chain-accuracy gate across $K$ consecutive training groups; otherwise the candidate is revised. The DRH is a heuristic specifying how to read the chain answer from the trace (\eg{} ``read the sign of the pitch component of the target orientation'').

\textbf{Training groups.} Each trace falls into one \emph{exploratory chain}, the action sequence the robot attempts during exploration; the per-task inventory is in Appendix~\ref{app:tasks}. Training traces are partitioned into groups keyed by (environment instance, exploratory chain), each group containing at most a fixed cap of environment instances and exploratory-chain types. At each iteration round, $K$ groups are sampled without replacement from this pool; once the pool is exhausted, the partition is re-randomized and sampling restarts. The policy keeps each round's training-trace distribution balanced across exploratory-chain variety, regardless of the total training data size.

\textbf{Commit gate.} A candidate clears commit when it passes a held-out chain-accuracy gate across $K$ consecutive training groups. The two hyperparameters $K$ and the gate threshold jointly trade off per-round wall-clock, agent-token consumption, training-trace coverage (larger $K$ samples more exploratory-chain variety and resists overfit to an easy subset), and the committed DRH's robustness (stricter gates filter out unreliable candidates at the cost of convergence speed). Specific values used in all reported runs are documented in \S\ref{sec:setup_main}.

\textbf{Iteration cost.} The wall-clock time and token consumption of the closed-loop iteration are not fixed properties of the pipeline: both grow with the amount of training data the agent reads per round, and both vary with the iteration agent itself, since different coding agents differ in tool-call strategy and context management. In our experiments, each task's iteration run converged in $1$--$5$ candidate-DRH rounds, with a typical round taking $5$--$30$ minutes of wall-clock and $50$k--$200$k tokens of agent context. The committed DRH itself is up to $1$k tokens, varying with task complexity and the iterations the agent runs, prepended once to the chain prompt at inference. Per-call inference cost is dominated by encoding the trace input, in particular the video frames.

\textbf{Inference.} A downstream model consumes the task-keyed DRH alongside the raw trace input. Our headline downstream model is a frozen base VLM (the \emph{Distilled-Prompt VLM}); the only inference-time difference from the \emph{Naked-Modality VLM} (same task, same model, no DRH) is the DRH itself. The same DRH also drives a second class of downstream model: programmatic baselines one-shot generated by a coding agent (\S\ref{sec:supervised_baselines}).

% \textbf{Controlled comparison.} To make the Distilled-Prompt vs.\ Naked-Modality comparison fair, we (i) fix the chain prompt and evaluation cap across both conditions in every (task, model) pair, (ii) keep the agent absent at inference (no tool calls), and (iii) require every DRH to clear a held-out chain-accuracy gate before commit, so the lift cannot come from overfit to the iteration set.

\subsection{Incremental multi-task via prompt-only artifacts}
\label{sec:multi_task}

Because the DRH is a single line per task, the system is incremental-multi-task: trained tasks share one base VLM under a task-keyed lookup, untrained tasks degrade to base-model behavior, and adding a task costs one closed-loop run.

%% file: sections/4_main_results.tex
% Section 4: Main Results (aggressively compressed; details to appendix)

We evaluate \emph{Closed-Loop Trace Distillation} along two axes. (Q1) Does the \emph{Distilled-Prompt VLM} lift chain accuracy over the \emph{Naked-Modality VLM} on the same frozen base? (Q2) Is the DRH precise enough to drive a programmatic implementation that matches direct LLM consumption of the same trace input?

\subsection{Experimental Setup}
\label{sec:setup_main}

\textbf{Models.} \emph{GPT-5.5} at maximum reasoning effort is the frozen base for both the Distilled-Prompt VLM and the Naked-Modality VLM, so the $\Delta$ is attributable to the DRH alone. \emph{HY-Embodied-0.5-X}~\citep{tencent2026hyembodied} is a recent embodied multimodal LLM run as a no-DRH reference.

\textbf{Tasks and Baselines.} We evaluate three simulator tasks (safe, lamp, door) and two real-robot tasks (bottle, cabinet); representative frames per task are in Figure~\ref{fig:task_gallery}. For every (task, base-model) pair, the Naked-Modality VLM is the best of $\{\text{video}, \text{proprio}, \text{video}+\text{proprio}\}$ conditions from the modality ablation (Table~\ref{tab:modality_ablation}); the chain prompt and evaluation cap are identical between the Naked-Modality and Distilled-Prompt VLMs. We report chain accuracy throughout. For GPT-5.5, we use maximum reasoning effort; for HY-Embodied, we use a $2048$-new-token cap on text-only conditions and a $4096$-new-token cap on video-bearing conditions, on a single H200 in bfloat16.

\textbf{Trace Collection.} Simulator traces are collected in IsaacGym~\citep{makoviychuk2021isaacgym} on an AdaManip~\citep{wang2025adamanip} task suite, with $10$ distinct environment instances per task varying in object structure and appearance. Training groups are keyed by (environment instance, exploratory chain) with a per-group cap of $7 \times 3 = 21$ episodes; each simulator task's data forms $5$ such groups, $1$ held out as evaluation ($n=21$) and $4$ forming the training pool sampled by the iteration loop. Real-robot traces are collected on physical hardware via teleoperation, with each task using a single setup and providing $60$ demonstrations ($3$ exploratory-chain types $\times$ $20$ episodes per chain); $15$ episodes per task ($5$ per chain) are reserved as held-out and shared between the Distilled-Prompt VLM and the modality ablation. The per-task exploratory-chain inventory is in Appendix~\ref{app:tasks}. Each episode ships with one fixed-camera video at $10$\,Hz (simulator: $128\times 128$ for safe, $512\times 512$ for lamp and door; real-robot: $640\times 480$) and a proprioceptive trajectory recording the 6-D target end-effector pose (Euler-angle rotation in degrees, Cartesian position in meters) and, when available, a 1-D gripper open/close state.

\textbf{Iteration training.} The closed-loop iteration uses $K=3$ and a chain-accuracy gate of $0.85$.

\subsection{Chain accuracy on simulator and real-robot tasks}
\label{sec:headline}

Table~\ref{tab:main_results} gives the headline result. The Distilled-Prompt VLM reaches at least $0.93$ on every task; the Naked-Modality VLM sits at $0.53$--$0.62$ on the same evaluation groups. The $+0.38$ to $+0.47$ absolute gap holds consistently across the three simulator tasks (safe, lamp, door) and the two real-robot tasks (bottle, cabinet), indicating that the DRH, not raw multimodal capacity, accounts for the lift. The five learned DRHs each encode a task-specific reading strategy over the proprioceptive trajectory; per-task DRH text is in Appendix~\ref{app:learned_prompts}.

\input{figures/table1_main_results.tex}

\subsection{Programmatic baselines from the DRH}
\label{sec:supervised_baselines}

We next test whether the DRH is precise enough to specify a non-VLM implementation. Given the DRH as the sole specification, a coding agent one-shot generates two supervised baselines on the same training groups. \emph{B1} is a fixed-grid scripted-rule learner that enumerates $153$ candidate rules (sign-of-delta, magnitude-threshold, extremum-position-bucket, and intermediate-extremum-vs-endpoint comparisons) on the proprioceptive features the DRH points to, and picks each rule's majority-vote training label. \emph{B2} is a pair of standard classifiers (logistic regression and random forest from scikit-learn~\citep{pedregosa2011sklearn}) trained on a 35-dim per-episode feature vector derived from the same features.

Table~\ref{tab:supervised_baselines} reports the held-out breakdown. On the simulator held-out split (three tasks, $n=63$), B1 and B2-LR both reach $63/63$, while B2-RF and the Distilled-Prompt VLM both reach $62/63$. The DRH alone is thus a sufficient specification: a coding agent with no other access to the iteration loop produces classifiers that match or exceed the frozen VLM consuming the same DRH.

% Per-task held-out denominators are normalized to $n=21$ for comparability; rates are approximate (rounded to the nearest integer numerator).
\begin{table}[t]
\centering
\small
\caption{Matched supervised baselines on the simulator held-out splits. B1 = fixed-grid scripted-rule learner (153 candidate rules); B2-LR / B2-RF = logistic regression / random forest on a 35-dim per-episode feature vector. Distilled-Prompt VLM copied from Table~\ref{tab:main_results}.}
\label{tab:supervised_baselines}
\begin{tabular}{lcccc}
\toprule
Task (held-out) & Distilled-Prompt VLM & B1 & B2-LR & B2-RF \\
\midrule
safe        & $21/21 = 1.000$ & $21/21 = 1.000$ & $21/21 = 1.000$ & $21/21 = 1.000$ \\
lamp        & $21/21 = 1.000$ & $21/21 = 1.000$ & $21/21 = 1.000$ & $21/21 = 1.000$ \\
door        & $20/21 \approx 0.952$ & $21/21 = 1.000$ & $21/21 = 1.000$ & $20/21 \approx 0.952$ \\
\midrule
\textbf{aggregate} & $62/63 \approx 0.984$ & $63/63 = 1.000$ & $63/63 = 1.000$ & $62/63 \approx 0.984$ \\
\bottomrule
\end{tabular}
\end{table}

%% file: figures/table1_main_results.tex
\begin{table}[t]
\centering
\small
\caption{Main results. The \emph{Naked-Modality VLM} is the best of \{video, proprio, video + proprio\} with no DRH on the same evaluation group (drawn from Table~\ref{tab:modality_ablation}); the \emph{Distilled-Prompt VLM} replaces only the DRH.}
\label{tab:main_results}
\vspace{2pt}
\begin{tabular}{lcccc}
\toprule
Task & Scene & Naked-Modality VLM & Distilled-Prompt VLM & $\Delta$ \\
\midrule
safe    & sim  & 12/21 = 0.571 \,(proprio)          & 21/21         & \textbf{+0.429} \\
lamp    & sim  & 13/21 = 0.619 \,(proprio)          & 21/21         & \textbf{+0.381} \\
door    & sim  & 12/21 = 0.571 \,(video + proprio)  & 20/21 = 0.952 & \textbf{+0.381} \\
bottle  & real & 8/15 = 0.533 \,(proprio)           & 14/15 = 0.933 & \textbf{+0.400} \\
cabinet & real & 8/15 = 0.533 \,(video + proprio)   & 15/15         & \textbf{+0.467} \\
\bottomrule
\end{tabular}
\end{table}

%% file: sections/5_ablations.tex
% Section 5: Modality Ablation
\label{sec:modality}

We strip the closed-loop step and ask whether either base model can answer the EMT-QA chain question from raw modalities alone. Each model runs three no-DRH modality conditions: video only, proprio only, and video + proprio, with the same chain prompt (Appendix~\ref{app:prompts}) and evaluation cap.

\input{figures/table2_modality_gpt55.tex}

Table~\ref{tab:modality_ablation} reports the three modality rows per (task, model) pair. HY-Embodied collapses to below chance on safe video + proprio ($1/21 = 0.048$, against a $1/3$ random baseline over safe's three candidate chains), and its video-only and proprio-only rows on door are pointwise identical ($7/21 = 0.333$), with only video + proprio breaking the tie. The best Naked-Modality VLM (GPT-5.5) reaches $0.571$--$0.619$ on the three simulator tasks and $0.533$ on both real-robot tasks, well below the Distilled-Prompt VLM's $0.933$--$1.000$. Neither model recovers the lift from raw modalities, locating the contribution in the DRH rather than the underlying modality stack.

\begin{figure}[t]
    \centering
    \includegraphics[width=0.6\linewidth]{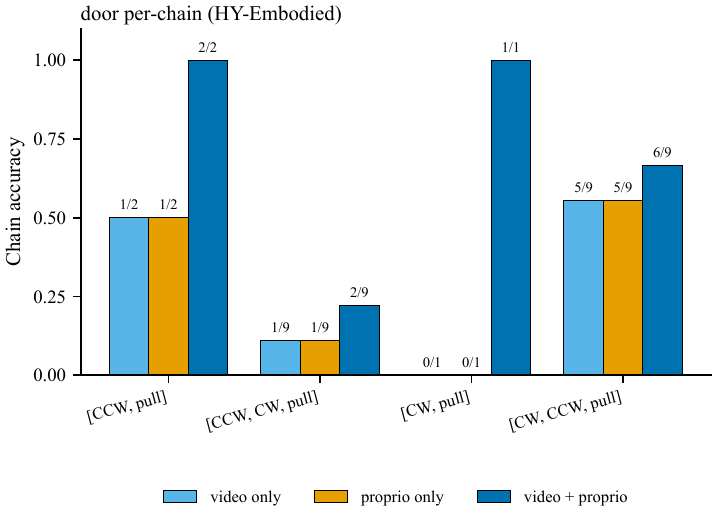}
    \caption{Door per-truth-chain accuracy for HY-Embodied-0.5-X. The video-only and proprio-only rows are pointwise identical across all four chains; both fall back to the chain-prompt prior. The video + proprio row breaks the tie on the two non-retry chains.}
    \label{fig:door_perchain}
\end{figure}

\paragraph{Per-truth-chain attribution on door.}
\label{app:perchain}
Figure~\ref{fig:door_perchain} breaks HY-Embodied's three modality rows down by the four ground-truth chains on door. The video-only and proprio-only rows are pointwise identical: [CCW, pull] $1/2$, [CCW, CW, pull] $1/9$, [CW, pull] $0/1$, [CW, CCW, pull] $5/9$; both fall back to the \emph{chain-prompt prior}, the predictions attainable from the chain prompt alone, without using the trace. The video + proprio row breaks the tie unevenly: perfect on the two non-retry chains, $2/9$ and $6/9$ on the two retry chains. The complementary-modality contribution is real but partial, far below the Distilled-Prompt VLM's $0.942$ aggregate.

\paragraph{What the DRH supplies that raw modalities do not.}
Taken together, the modality ablation and the per-truth-chain attribution point to a specific gap. The base VLM is not bottlenecked on modality bandwidth: video, proprioception, and their combination all fall in the same $0.53$--$0.62$ band on the same frozen model, and on door three of HY-Embodied's modality rows partition the four ground-truth chains as if drawn from the chain-prompt prior. What the VLM lacks under the Naked-Modality condition is a task-specific reading procedure: across an exploratory trace it can sample frames and proprioceptive values, but it cannot decide which dimension of the trace carries the precondition, whether to read its sign, its extremum, its phase-split structure, or its gripper-segmented composition, and how to map the chosen reading to an option in the chain prompt. The DRH supplies exactly this missing layer: a one-line, task-specific natural-language pointer that names the procedure rather than the answer, leaving the perception and the chain-prompt selection to the same frozen base model. This is consistent with the observation that the same DRH transfers without modification to a coding agent (\S\ref{sec:supervised_baselines}), which inherits the procedure but not the VLM: the artifact carries the task-level readout, not VLM-specific conditioning.

%% file: figures/table2_modality_gpt55.tex
\begin{table}[t]
\centering
\small
\caption{Modality ablation under no DRH. Each cell reports the number of correctly predicted episodes over the total, followed by chain accuracy in parentheses. \textbf{Bold} marks the best modality per task. The best modality is non-monotone across tasks.}
\label{tab:modality_ablation}
\vspace{2pt}
\begin{tabular}{llccc}
\toprule
Task & Scene & video & proprio & video + proprio \\
\midrule
\multicolumn{5}{l}{\emph{GPT-5.5 (frozen base VLM)}} \\
lamp    & sim  & 6/21 (0.286) & \textbf{13/21 (0.619)} & \textbf{13/21 (0.619)} \\
safe    & sim  & 6/21 (0.286)  & \textbf{12/21 (0.571)} & 11/21 (0.524) \\
door    & sim  & 11/21 (0.524) & 11/21 (0.524)          & \textbf{12/21 (0.571)} \\
bottle  & real & 6/15 (0.400)  & \textbf{8/15 (0.533)}  & 5/15 (0.333) \\
cabinet & real & 7/15 (0.467)  & 6/15 (0.400)           & \textbf{8/15 (0.533)} \\
\midrule
\multicolumn{5}{l}{\emph{HY-Embodied-0.5-X}} \\
lamp    & sim  & 8/21 (0.381)  & \textbf{10/21 (0.476)} & 8/21 (0.381) \\
safe    & sim  & 7/21 (0.333)  & \textbf{8/21 (0.381)}  & 1/21 (0.048) \\
door    & sim  & 7/21 (0.333)  & 7/21 (0.333)           & \textbf{11/21 (0.524)} \\
\bottomrule
\end{tabular}
\end{table}

%% file: sections/6_limitations.tex
% Section 6: Limitations (mandatory at CoRL; framework-scope only)

\textbf{Assumptions.} Our pipeline assumes ground-truth chain labels at training time and a chain answer expressible by a one-line heuristic over the trace. Across our five tasks the discovered heuristics all read the proprioceptive trajectory (sign-of-delta, extremum-vs-endpoint, gripper-segmented $\Delta$, or combinations); tasks whose discriminator lies primarily in the visual channel, tasks beyond this empirical grammar, and label-free regimes are out of scope.
% Future work could extend the heuristic grammar to multi-column conjunctions or temporal-window features, and replace the ground-truth-label requirement with a weakly-supervised iteration loop driven by partial chain agreement.

% , and we do not benchmark against task-specific supervised methods , and we do not benchmark against task-specific supervised methods
\textbf{Trace coverage.} All evaluation traces come from two clean sources: scripted simulator traces and teleoperated real-robot demonstrations, with every trace recording a full successful task completion. Future work could extend the evaluation to partial-completion traces (a trace that ends after the first failed door-pull, for instance, should still let the model infer that unlocking precedes pulling) and to noisier sources such as policy rollouts on the same tasks.

\textbf{Generalization.} Across three simulator tasks and two real-robot tasks on distinct embodiments, the DRH transfers along both the task and embodiment axes. However, all five tasks are single-arm, stationary-base, single-object, and short-horizon; bimanual coordination, mobile manipulation, multi-object scenes, and longer-horizon chains remain unevaluated. Future work could extend the suite along these axes and characterize whether one-line heuristic discovery still suffices.

%% file: sections/7_conclusion.tex
% Section 7: Conclusion

We presented \emph{Closed-Loop Trace Distillation}, a pipeline for EMT-QA. The pipeline lifted chain accuracy of the Distilled-Prompt VLM by $+0.38$ to $+0.47$ over the best Naked-Modality VLM across three simulator and two real-robot tasks. Handed the same DRH as the sole specification, a coding agent one-shot generated programmatic baselines that matched direct LLM consumption on the simulator held-out split. Because the DRH is consumed identically by a frozen base VLM and by a coding-agent-generated program, it decouples the trained artifact from the model that reads it: training-time search produces a single human-auditable line of natural language, and any inference-time consumer able to read the trace inherits the task-level readout without weight updates or agent calls. We hope this dual-consumer, prompt-resident view of multimodal trace reasoning will encourage future work that treats the reading procedure, rather than the policy weights, as the portable unit of transfer across embodiments and downstream consumers.

%% file: sections/A_appendix.tex
% Appendix A--E

% Per-group K=3 breakdown and Wilson 95\% CI section commented out for current submission.
% If reviewers ask about per-group splits or small-sample CIs, restore from git history.

\section{Evaluation tasks}
\label{app:tasks}

\begin{figure}[h]
    \centering
    \includegraphics[width=0.85\linewidth]{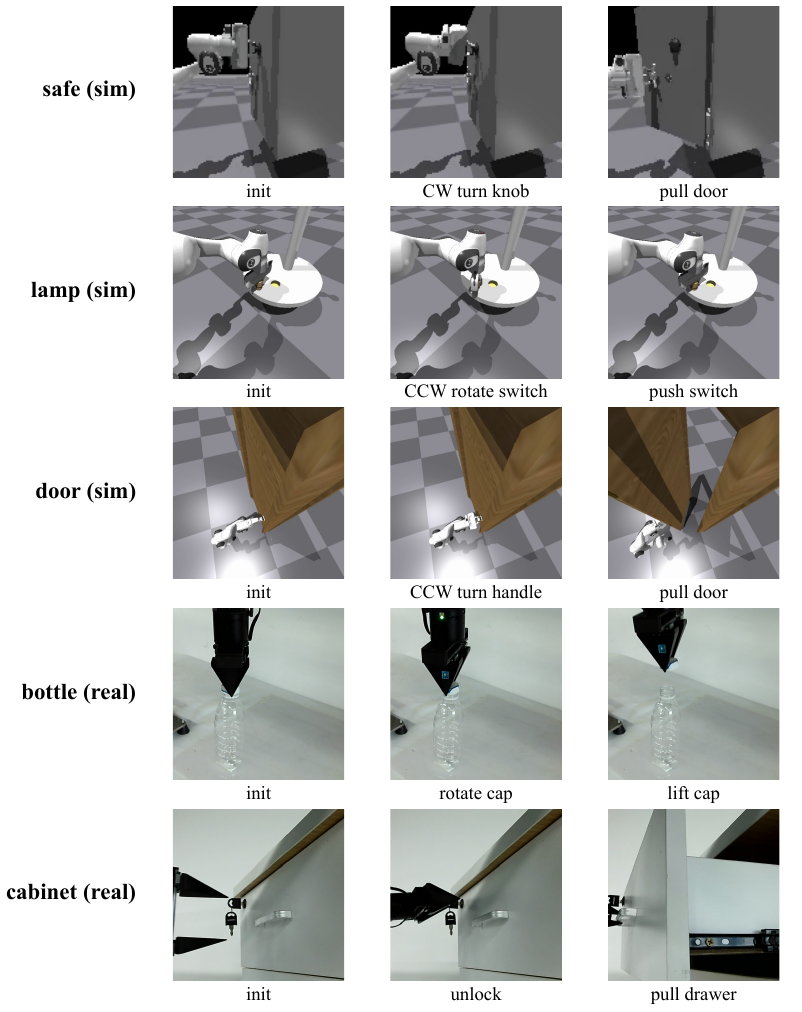}
    \caption{Representative frames for the five evaluation tasks. Each row corresponds to one task; the leftmost frame is the initial state and subsequent frames show one keyframe per action in the minimal-success chain. Frames are rescaled and center-cropped to a common aspect ratio for layout only.}
    \label{fig:task_gallery}
\end{figure}

Figure~\ref{fig:task_gallery} shows the five evaluation tasks introduced in \S\ref{sec:setup_main}: the three simulator tasks (safe, lamp, door) followed by the two real-robot tasks (bottle, cabinet). Each row visualizes the same per-episode video stream the frozen base VLMs in \S\ref{sec:main_results} and \S\ref{sec:modality} consume alongside the proprioceptive trajectory. The displayed frames are representative keyframes selected for visual clarity; their left-to-right order is illustrative and does not encode the precise temporal sequence of execution.

% The frozen-base-VLM rows uniformly cap the video token budget at $12$ frames per episode (\texttt{frame\_max\_count}\,=\,12) via uniform sub-sampling along the episode timeline; the trajectory remains full-length and is not sub-sampled.

\paragraph{Per-task exploratory-chain inventory.}
The data for each task is partitioned into the following exploratory-chain types, the distinct trajectory categories under which the demonstrations are generated. \emph{Retry chains} are traces in which the robot first attempts a wrong direction or contact, then commits to the correct action; the minimal-success chain (the prediction target) collapses retry traces by writing only the final effective action sequence (for cabinet, by preserving the unlock step recovered between two pull attempts).
\begin{itemize}
    \item \textbf{safe} (3 types): [pull-door], [CCW turn-knob, pull-door], [CCW turn-knob, CW turn-knob, pull-door] (the third is a retry-then-succeed chain).
    \item \textbf{lamp} (5 types): [push-switch], [push-switch, CCW rotate-switch], [push-switch, CCW rotate-switch, CW rotate-switch], [push-switch, CW rotate-switch], [push-switch, CW rotate-switch, CCW rotate-switch] (the three- and two-step variants include retry-then-succeed patterns in which the last rotation is the final effective direction).
    \item \textbf{door} (one-go variant, 4 types): [CCW turn-handle, pull-door], [CW turn-handle, pull-door], [CCW turn-handle, CW turn-handle, pull-door], [CW turn-handle, CCW turn-handle, pull-door]; about $56\%$ of the $160$ episodes fall in the two retry chains.
    \item \textbf{bottle} (real robot, 3 types): [lift-cap], [rotate-cap, lift-cap], [lift-cap, rotate-cap, lift-cap] (the third is a retry-then-succeed chain in which the robot attempts a lift, fails, then commits to rotate-then-lift).
    \item \textbf{cabinet} (real robot, 3 types): [pull-drawer], [unlock, pull-drawer], [pull-drawer, unlock, pull-drawer] (the third is a retry-then-succeed chain in which the robot attempts pull-drawer, encounters resistance, then performs unlock between two pull-drawer attempts).
\end{itemize}

\section{Per-task learned DRH text}
\label{app:learned_prompts}

The trained artifact this paper produces is one DRH string per task. The reading strategy for each task is summarized below.

\paragraph{safe.} Read the \emph{sign} of the change in the pitch component of the target end-effector orientation between trajectory start and end. Positive sign $\Rightarrow$ counter-clockwise (CCW) turn-knob; negative sign $\Rightarrow$ clockwise (CW) turn-knob. Output chain is [$\langle$rotation direction$\rangle$, pull-door].
\paragraph{lamp.} If the trajectory exhibits a push-switch morphology (single short translation, no sustained rotation), output [push-switch]; otherwise read the sign of the change in the yaw component of the target end-effector orientation. Positive $\Rightarrow$ CCW rotate-switch; negative $\Rightarrow$ CW rotate-switch. Retry-chain net-change handled by a tiebreaker on absolute pitch extremum.
\paragraph{door.} Phase-split the trajectory at the first sign-change of the pitch-component delta; compare the pitch extremum against both endpoints; if the extremum is at an intermediate step the chain is a retry chain ([CCW, CW, pull] or [CW, CCW, pull]). Otherwise the chain is a non-retry single-direction chain.
\paragraph{bottle (real robot).} Read the sign of a single proprioceptive column to discriminate whether a rotation phase precedes the lift. Output chain is [lift-cap] when no rotation phase is detected; otherwise [rotate-cap, lift-cap].
\paragraph{cabinet (real robot).} Use gripper open/close events to split the trajectory into contact segments; keep only segments with the gripper closed. Per-segment median comparison: a change in target end-effector orientation $\gtrsim 30^\circ$ on any axis OR a change in target end-effector position $\gtrsim 3$--$5$ cm flags distinct action types. Within-segment kinematics: drawer-pull (linear translation, no rotation) vs.\ lock-open (rotation only on one axis, negligible translation). Output chain includes both actions when the per-segment medians match the two signatures; a lock-open sandwiched between two drawer-pulls is treated as the corrective action that satisfies the latent precondition (cabinet locked) revealed by the first failed pull, rather than discarded as a spurious between-pull deviation.

\section{Inference prompt templates}
\label{app:prompts}

This section documents the prompt templates used at inference time across the three modality conditions evaluated in the modality ablation (\S\ref{sec:modality}) and the Distilled-Prompt VLM (\S\ref{sec:main_results}). All prompts use the \texttt{structured} prompt style. The actual prompts are in Chinese; faithful English translations are given below. Placeholders are shown in \texttt{\{braces\}}.

\subsection{Chain-prediction prompt (modality-dependent)}

The chain-prediction prompt is the primary instruction sent to the model. It varies by input modality:

\paragraph{Video-only.}
\begin{quote}\small
\{media\_prefix\} demonstrates an attempt at the robotic arm task `\{command\}'. Based only on the input video frames, select the step sequence most consistent with the actual process: \{choices\}. First internally check the actual contact objects, task-critical state changes, and action ordering.
\end{quote}

\paragraph{Proprio-only.}
\begin{quote}\small
\{media\_prefix\} records an attempt at the robotic arm task `\{command\}'. Based only on the input robot action trajectory, select the step sequence most consistent with the actual process: \{choices\}. First internally check the contact segments, action types, and action ordering reflected by position, rotation, and gripper changes in the trajectory; do not supplement steps not observed in the trajectory based on other priors or common sense.
\end{quote}

\paragraph{Video + proprio.}
\begin{quote}\small
\{media\_prefix\} demonstrates an attempt at the robotic arm task `\{command\}'. Based on the input video frames and the optional robot action trajectory, select the step sequence most consistent with the actual process: \{choices\}. First internally check the actual contact objects, task-critical state changes, action ordering, and whether position, rotation, and gripper changes in the trajectory support the phase judgment.
\end{quote}

\noindent In all three variants, the prompt ends with the answer-format instruction and, when the Distilled-Prompt VLM condition is active, the task-keyed DRH:

\begin{quote}\small
[Task-specific judgment requirement: \{additional\_prompt\}]\quad Place the answer digit on the first line; output only one selectable digit on the first line without preceding explanation; from the second line onward you may briefly state reasoning.
\end{quote}

\noindent The bracketed DRH block is present only in the Distilled-Prompt VLM condition; the Naked-Modality VLM omits it.

\subsection{Proprio context instruction (proprio-bearing conditions)}

When the input includes proprio data (proprio-only or video+proprio), a proprio-context instruction block is appended after the chain-prediction prompt. It consists of (i) a modality-role statement, (ii) a noise caveat, and (iii) the proprio JSON payload.

\paragraph{Proprio-only instruction.}
\begin{quote}\small
The robot action trajectory is as follows. The trajectory data contains only action commands; it does not contain environment observations, object states, phase labels, success flags, or correct answers. \{noise\_caveat\} Synthesize position, rotation, gripper, and other signals to roughly infer contact/movement phases and their ordering, then match the inferred phase sequence to the available options; if a candidate holds only under subtle trajectory differences that are not significant within the noise range, prefer the simpler option.
\end{quote}

\paragraph{Video + proprio instruction.}
\begin{quote}\small
The raw robot action trajectory is appended below. The trajectory data contains only action commands; it does not contain environment observations, object states, phase labels, success flags, or correct answers. \textbf{Use video as primary evidence and trajectory as auxiliary}: first use contact objects and state changes visible in the video as the main evidence, then use position, rotation, and gripper changes in the trajectory to verify possible phases. \{noise\_caveat\} Only match phase sequences that video and trajectory \textbf{roughly jointly support} to the available options; do not name phases based solely on trajectory numbers.
\end{quote}

\paragraph{Noise caveat (shared).}
\begin{quote}\small
\textbf{Note: trajectory data may come from model inference and contain noise}---the gripper may randomly open/close during approach, joints may jitter, and the arm may attempt the same action, fail, and retry. Therefore \textbf{do not strongly rely on gripper state to segment phases}, and do not treat any single signal in the trajectory as a precise phase boundary; the trajectory can only serve as a \textbf{coarse} auxiliary reference.
\end{quote}

\subsection{Proprio data payload}

The proprio JSON payload follows the instruction block and is formatted as a fenced code block. Under the \texttt{raw} representation (used in all reported runs), each row contains a timestep index and the verbatim action vector:

\begin{quote}\small\ttfamily
\{"t": 0, "action": [0.12, -0.03, 0.41, ...]\}\\
\{"t": 1, "action": [0.12, -0.02, 0.41, ...]\}\\
...
\end{quote}

\noindent The schema header declares the action dimensionality and column semantics (position, rotation-6D, gripper when present). No derived features, phase labels, or success flags are included in the payload.

\subsection{Task-specific placeholders}

Table~\ref{tab:prompt_placeholders} lists the per-task values substituted into the template placeholders.

\begin{table}[h]
\centering\small
\caption{Per-task placeholder values used in the inference prompt templates.}
\label{tab:prompt_placeholders}
\begin{tabular}{@{}lll@{}}
\toprule
Task & \texttt{\{command\}} & \texttt{\{choices\}} \\
\midrule
safe & open safe & 1.~pull-door~~2.~CW turn-knob$\to$pull-door~~3.~CCW turn-knob$\to$pull-door \\
lamp & turn on lamp & 1.~push-switch~~2.~CW rotate-switch~~3.~CCW rotate-switch \\
door & open door & 1.~CW turn-handle$\to$pull-door~~2.~CCW turn-handle$\to$pull-door \\
bottle & open bottle & 1.~lift-cap~~2.~rotate-cap$\to$lift-cap \\
cabinet & open drawer & 1.~pull-drawer~~2.~unlock$\to$pull-drawer \\
\bottomrule
\end{tabular}
\end{table}